\documentclass[conference]{IEEEtran}
\IEEEoverridecommandlockouts

\usepackage{tcolorbox}
\usepackage{hyperref} 
\usepackage{adjustbox}
\usepackage{cite}
\usepackage{amsmath,amssymb,amsfonts}
\usepackage{algorithmic}
\usepackage{graphicx}
\usepackage{textcomp}
\usepackage{xcolor}
\def\BibTeX{{\rm B\kern-.05em{\sc i\kern-.025em b}\kern-.08em
    T\kern-.1667em\lower.7ex\hbox{E}\kern-.125emX}}

\begin{document}
\title{Medical Vision Language Models as Policies for Robotic Surgery\\
{\footnotesize \textsuperscript{}}
\thanks{}
}

\author{\IEEEauthorblockN{\textsuperscript{} Akshay Muppidi}
\IEEEauthorblockA{\textit{Department of Computer Science} \\
\textit{Stony Brook University}\\
Stony Brook, USA \\
akshaym0815@gmail.com}
\and
\IEEEauthorblockN{\textsuperscript{} Martin Radfar}
\IEEEauthorblockA{\textit{Department of Computer Science} \\
\textit{Stony Brook University}\\
Stony Brook, USA \\
radfar@cs.stonybrook.edu}}

\maketitle

\begin{abstract}
Vision-based Proximal Policy Optimization (PPO) struggles with visual observation-based robotic laparoscopic surgical tasks due to the high-dimensional nature of visual input, the sparsity of rewards in surgical environments, and the difficulty of extracting task-relevant features from raw visual data. We introduce a simple approach integrating MedFlamingo, a medical domain-specific Vision-Language Model, with PPO. Our method is evaluated on five diverse laparoscopic surgery task environments in LapGym, using only endoscopic visual observations. MedFlamingo PPO  outperforms and converges faster compared to both standard vision-based PPO and OpenFlamingo PPO baselines, achieving task success rates exceeding 70\% across all environments, with improvements ranging from 66.67\% to 1114.29\% compared to baseline. By processing task observations and instructions once per episode to generate high-level planning tokens, our method efficiently combines medical expertise with real-time visual feedback. Our results highlight the value of specialized medical knowledge in robotic surgical planning and decision-making.

\end{abstract}


\section{Introduction}
Robot-assisted laparoscopic surgery stands as a significant advancement in minimally invasive medical procedures \cite{csirzo2024robot, roh2018robot, ruurda2002feasibility, rodriguez2016laparoscopic}. This approach enables surgeons to manipulate robotic instruments via a console, achieving precise maneuvers within the patient's body while minimizing trauma \cite{Zong2022-oz}. The potential for enhancing these procedures through autonomous systems has sparked significant interest in the medical and artificial intelligence communities.

Reinforcement learning (RL) \cite{kaelbling1996reinforcementlearningsurvey} has emerged as a promising paradigm for training autonomous systems \cite{mahmood2018benchmarking}. In RL, algorithms learn decision-making strategies by interacting with an environment, aiming to maximize cumulative rewards. Recent developments in this field include LapGym \cite{scheikl2023lapgym}, a suite of RL environments designed to simulate common laparoscopic surgical training tasks. These environments, inspired by the Fundamentals of Laparoscopic Surgery (FLS) program \cite{peters2004development, khajeh2023outcomes}, encompass crucial skills such as spatial reasoning, tissue manipulation, precise cutting, and thread manipulation. LapGym's initial results demonstrated that state-of-the-art RL algorithms \cite{schulman2017trustregionpolicyoptimization, chen2021decisiontransformerreinforcementlearning, attia2018globaloverviewimitationlearning, wang2016duelingnetworkarchitecturesdeep, mnih2013playingatarideepreinforcement, haarnoja2018softactorcriticoffpolicymaximum} like PPO \cite{articlePPO} could learn these skills effectively when provided with meticulously crafted state observations. However, when faced with endoscopic visual observations—akin to what would be available in real-world scenarios—these algorithms struggled, failing to achieve success rates beyond 30\% for most tasks.

Concurrently, recent advancements in VLMs, such as RT-2 \cite{brohan2023rt} and OpenVLA \cite{kim2024openvla}, have significantly impacted robotic learning and control. These models, leveraging large-scale pretraining on diverse internet data, have been successfully employed for robotic representation learning, task planning, and execution in modular systems. Of particular relevance to our work is using VLMs to generate high-level action tokens or representations that enhance existing reinforcement learning algorithms. For instance, \cite{szot2024largelanguagemodelsgeneralizable, wang2023voyageropenendedembodiedagent,chen2024visionlanguagemodelsprovidepromptable} have shown that large language models can be adapted into generalizable policies for embodied and agent tasks. 

Our research builds upon these findings, specifically focusing on the medical domain. We utilize a pre-trained Medical VLM as a backbone for PPO in robotic surgical policy learning. This simple integration allows for significant improvements over baseline PPO across various tasks, including dissection, grasping, thread manipulation, and spatial reasoning. Notably, our approach achieves success rates exceeding 70\% across all tasks, with some tasks reaching up to 95\% success rate.

By utilizing image-based endoscopic observations, our method closes the performance gap in simulation for automated robotic surgical tasks and leads the way toward a substantial healthcare impact for more precise, autonomous, real-world robotic surgery.

\section{Method}

\subsection{MedFlamingo}

Our method builds upon the integration of a vision-language model (VLM) and reinforcement learning, leveraging domain-specific medical expertise to enhance robotic policy learning. Following approaches such as \cite{szot2024largelanguagemodelsgeneralizable, hazra2023egotv}, which utilize egocentric observations and language instructions for task execution, we incorporate endoscopic vision observations alongside language instructions. This combination enables the agent to process surgical tasks in a manner informed by both visual feedback and textual task descriptions.

\begin{figure*}[htbp]
\centerline{\includegraphics[width=0.7\linewidth]{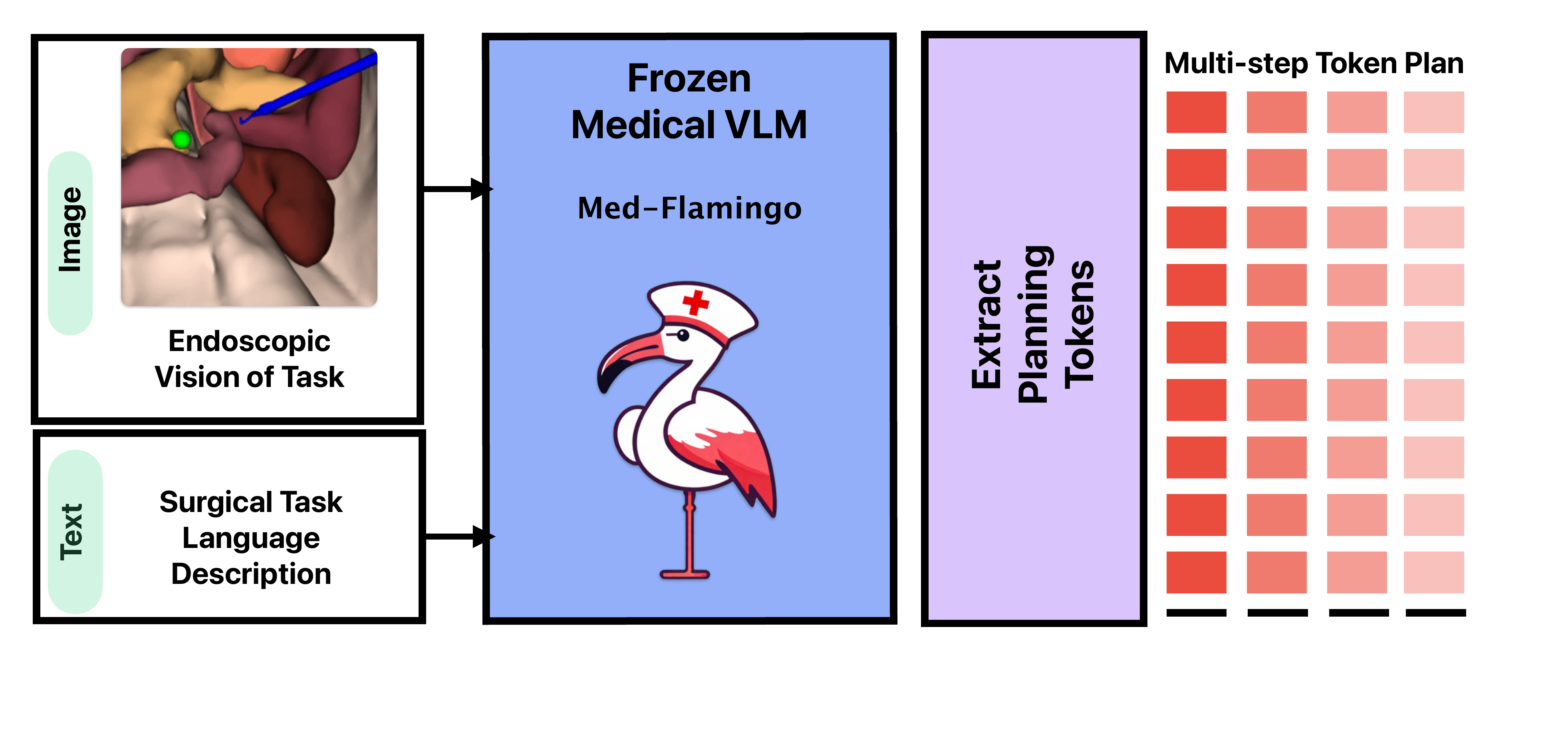}}
\caption{MedFlamingo processes the initial observation and task description to generate planning tokens. These tokens can be concatenated with ResNet-encoded visual features at each timestep.}
\label{tokens}
\end{figure*}

To address the computational overhead of large VLMs, we optimize the token generation process by invoking MedFlamingo only once at the start of each episode. This design choice avoids the need for repeated inference, reducing computational complexity while retaining high-level task-specific guidance. MedFlamingo processes the initial environment observation and task description to generate high-level planning tokens, denoted as \( m_t \in \mathbb{R}^k \) (Figure \ref{tokens}). These tokens encapsulate the domain-specific semantic information required for policy optimization.

The generated tokens are combined with visual features extracted from a ResNet encoder. The encoder processes endoscopic RGB observations, producing visual embeddings \( o_t \in \mathbb{R}^n \). These embeddings capture spatial and structural details of the surgical environment. The concatenated feature vector \( s_t = [m_t; o_t] \in \mathbb{R}^{k+n} \) serves as the input to the Actor and Critic networks, ensuring that both visual and textual information contribute to the decision-making process.

MedFlamingo, a multimodal (7B parameters) few-shot learner adapted for the medical domain, was selected as the VLM due to its specialized pretraining on medical data \cite{moor2023med}. This adaptation allows MedFlamingo to interpret surgical scenes and instructions with greater precision than general-purpose models. The model is frozen during training, ensuring consistent planning token generation across episodes.

\subsection{Online PPO}

The reinforcement learning component of our method employs Proximal Policy Optimization (PPO) \cite{articlePPO}, a robust algorithm for policy gradient optimization. PPO is well-suited for continuous action spaces and offers stable convergence through its clipped surrogate loss.

The state representation \( s_t \) combines MedFlamingo's high-level planning tokens \( m_t \) with ResNet-encoded visual features \( o_t \). This dual-modality input provides the Actor and Critic networks with a rich representation of the environment, balancing task-specific planning with detailed visual observations.

The PPO algorithm updates the Actor and Critic networks based on the clipped surrogate loss:
\begin{equation}
    L^{CLIP}(\theta) = \hat{\mathbb{E}}_t\left[\min(r_t(\theta)\hat{A}_t, \text{clip}(r_t(\theta), 1-\epsilon, 1+\epsilon)\hat{A}_t)\right]
\end{equation}
where \( r_t(\theta) = \frac{\pi_\theta(a_t|s_t)}{\pi_{\theta_{old}}(a_t|s_t)} \) is the probability ratio between the new and old policies, \( \hat{A}_t \) is the estimated advantage, and \( \epsilon \) is a hyperparameter controlling the trust region. The total loss is defined as:
\begin{equation}
    L^{TOTAL}(\theta) = \hat{\mathbb{E}}_t\left[L^{CLIP}(\theta) - c_1 L^{VF}(\theta) + c_2 S[\pi_\theta](s_t)\right]
\end{equation}
where \( L^{VF} \) is the value function loss, \( S \) is the entropy bonus encouraging exploration, and \( c_1, c_2 \) are weighting coefficients.

The Actor network \( \pi_\theta: \mathbb{R}^{k+n} \rightarrow \mathbb{R}^{|A|} \) maps the concatenated input \( s_t \) to action probabilities or parameters of a continuous distribution over the action space \( A \). The Critic network \( V_\phi: \mathbb{R}^{k+n} \rightarrow \mathbb{R} \) estimates the value function, guiding the policy updates to maximize cumulative rewards.

For comparison, two baseline PPO variants are implemented:
\begin{itemize}
    \item \textbf{Vision-based PPO}: Uses ResNet features \( o_t \) without planning tokens \( m_t \).
    \item \textbf{OpenFlamingo PPO} \cite{awadalla2023openflamingoopensourceframeworktraining}: Utilizes planning tokens generated by a general-purpose VLM, OpenFlamingo, instead of the domain-specific MedFlamingo.
\end{itemize}

The agent interacts with the LapGym environment at each timestep \( t \), receiving an observation and reward while generating an action \( a_t \). The reward signal updates the Actor and Critic networks using the PPO update rule. Training is performed over multiple episodes to evaluate the effectiveness of MedFlamingo's tokens compared to baselines.

\begin{figure*}[htbp]
\centerline{\includegraphics[width=0.7\linewidth]{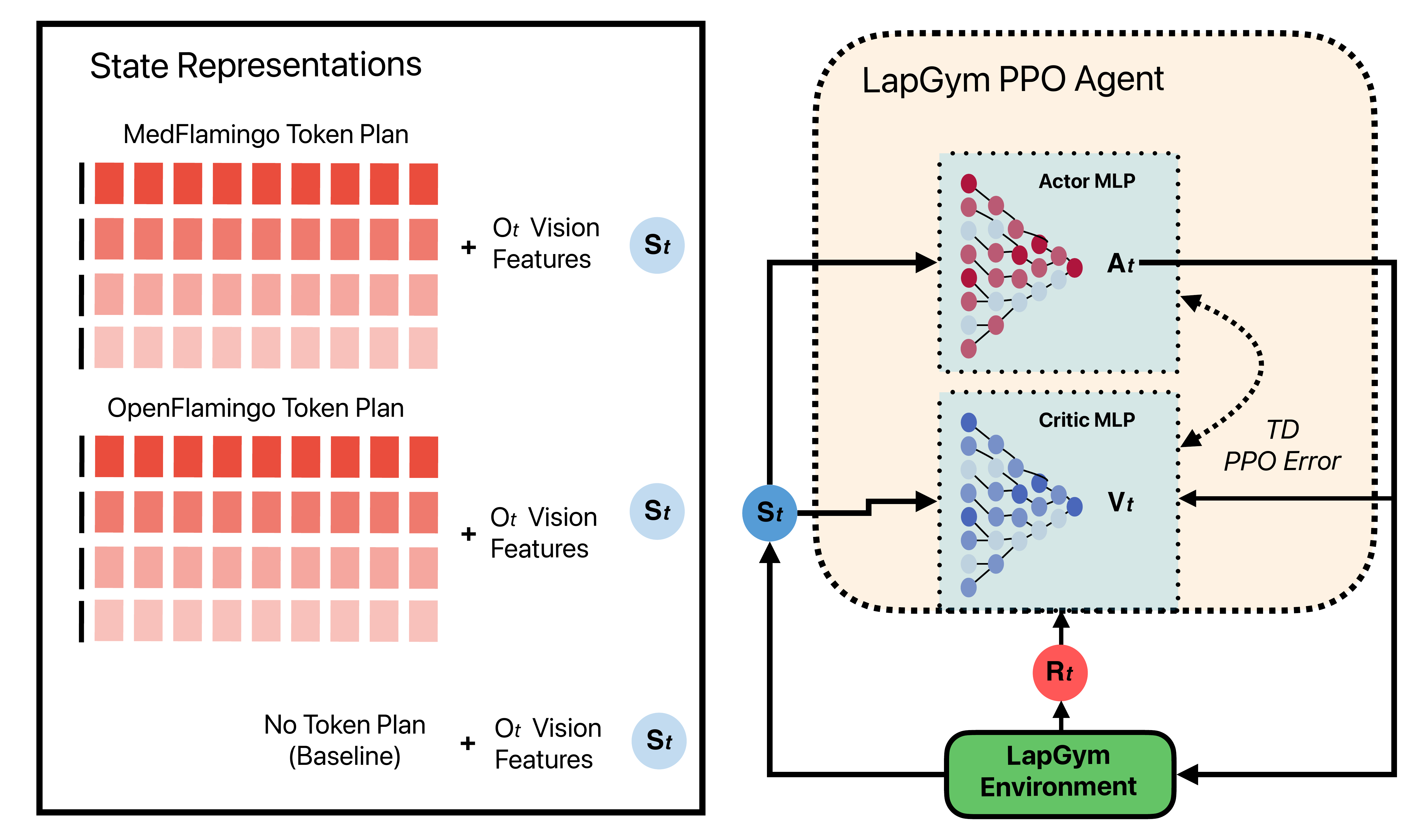}}
\caption{Illustration of the PPO agent's interaction with the LapGym environment. At each time step \(t\), the agent observes the environment state and uses its Critic and Actor networks to produce a value estimate \(v_t\) and an action \(a_t\), respectively. The action \(a_t\) is applied to the environment, which responds with the next state and a reward signal. This reward is used to update both the Actor and Critic networks through the PPO update mechanism. The environment state representation combines ResNet-encoded visual features with optional task-relevant planning tokens: MedFlamingo tokens (domain-specific), OpenFlamingo tokens (general-purpose), or no tokens (Baseline PPO).}
\label{state}
\end{figure*}

Figure \ref{tokens} illustrates the generation of planning tokens by MedFlamingo. The integration of tokens with ResNet features ensures that the agent's policy decisions leverage both high-level guidance and detailed visual information. Figure \ref{state} details the interaction loop between the PPO agent and the environment, highlighting the role of MedFlamingo, OpenFlamingo, and baseline representations in state encoding and action generation.

\section{Results}

We evaluate MedFlamingo PPO alongside two baseline methods \cite{articlePPO, awadalla2023openflamingoopensourceframeworktraining} across five diverse environments in LapGym. Figure 1 illustrates the performance of all three approaches on these tasks. All environments have endoscopic RGB-based camera observations provided by the LapGym library, simulating realistic visual input for laparoscopic procedures. Our baselines include standard PPO and OpenFlamingo \cite{awadalla2023openflamingoopensourceframeworktraining} PPO, with the latter using the same VLM architecture as MedFlamingo PPO but lacking medical domain knowledge fine-tuning.

\subsection{Environments}
\label{sec:experiments}

\subsubsection{DeflectSpheresEnv (Spatial Reasoning Track)}

The DeflectSpheresEnv is designed to test spatial awareness and precise instrument handling in a simulated surgical setting. This environment features a flat board with spheres on flexible stalks. Two electrocautery hooks (blue and red) operate in TPSD (Tool Position, Speed, and Direction) space to deflect the active, color-matched sphere while avoiding others. Successful deflections turn spheres green, and the task is complete when all required spheres are deflected. This environment emphasizes spatial reasoning under motion constraints imposed by RCM (Remote Center of Motion) \cite{nasiri2024admittance}, mimicking challenges in laparoscopic surgery.

\subsubsection{TissueManipulationEnv (Deformable Object Manipulation and Grasping Track)}

This environment simulates delicate manipulation of deformable tissues, a critical skill in laparoscopic procedures. The scenario involves a deformable yellow tissue (representing a gallbladder) attached to a rigid red structure (simulating a liver). The agent uses a laparoscopic grasper, controlled in Cartesian coordinates, to guide a visual landmark (a black dot) on the tissue to a target location (a blue dot). Success is achieved when the landmark is within 2 mm of the target, with the additional complexity of randomized landmark placement. This environment emphasizes learning the dynamic relationship between the grasper's actions and the deformation of the tissue.

\subsubsection{RopeCuttingEnv (Dissection Track)}

The RopeCuttingEnv focuses on precision dissection, a critical aspect of many surgical procedures. This environment features deformable ropes stretched between two walls. Using an electrocautery hook controlled in TPSD space, the agent must navigate to and cut a highlighted green rope. After a successful cut, a new rope is randomly highlighted. The task is completed when the agent cuts the required number of correct ropes. Precision is critical, as cutting incorrect ropes leads to task failure, mimicking the real-world consequences of surgical errors such as damaging unintended tissues.

\subsubsection{RopeThreadingEnv (Thread Manipulation Track)}

The RopeThreadingEnv tests fine motor control, bimanual dexterity, and spatial planning. This environment consists of a long thread, a set of eyelet screws arranged on a board, and two laparoscopic graspers controlled in TPSD space. The agent must guide the thread through the eyelets in a specific sequence and direction. Success is achieved when the thread passes through all eyelets in the correct order. This task simulates the challenges of threading or suturing in surgical contexts, where precise control of the thread's shape and tension is required.

\subsubsection{PickAndPlaceEnv (Deformable Object Manipulation and Grasping Track)}

The PickAndPlaceEnv tests an agent's ability to handle deformable objects with fine precision. This environment features a 3×3 grid of pegs, a deformable torus, and a laparoscopic grasper controlled in TPSD space and jaw angle. The agent must grasp the torus, lift it to a specified height, and place it onto the peg matching its color. Success is achieved when both the picking and placing phases are completed correctly. The environment allows parameterization of the torus's deformability and stiffness, adding complexity to the task. This setup mimics real-world challenges in manipulating soft tissues or highly mobile structures in surgical procedures.

\begin{figure*}[htbp]
\centering
\includegraphics[width=1.01\textwidth]{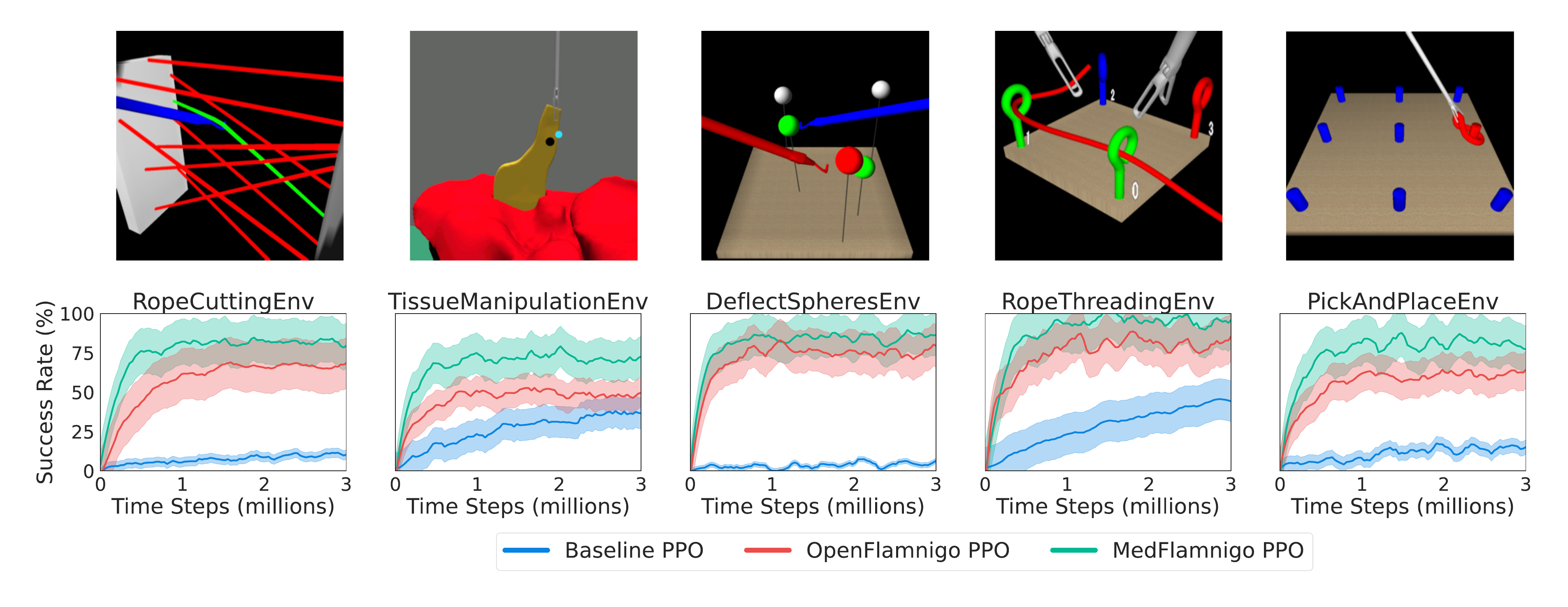}
\caption{Performance comparison (10 seed runs for each method) of MedFlamingo PPO, OpenFlamingo PPO, and Baseline PPO on the RopeCuttingEnv, TissueManipulationEnv, DeflectSpheresEnv, RopeThreadingEnv, and PickAndPlaceEnv. Success rates are reported on test environment seeds and normalized over total success rates.}
\label{fig}
\end{figure*}

\subsection{MedFlamingo PPO Improves Task Success Rates}

Quantitative analysis reveals substantial improvements in normalized success rates when comparing MedFlamingo PPO to Baseline PPO across test environment seeds. In the RopeCuttingEnv (Dissection Track), MedFlamingo PPO achieved a \textbf{256.52\%} increase in performance. The TissueManipulationEnv (Deformable Object Manipulation and Grasping Track) saw a \textbf{77.50\%} improvement. Most notably, the DeflectSpheresEnv (Spatial Reasoning Track) demonstrated an exceptional \textbf{1114.29\%} increase, highlighting MedFlamingo PPO's strength in complex spatial reasoning tasks. The PickAndPlaceEnv (Deformable Object Manipulation and Grasping Track) saw a \textbf{207.69\%} improvement, showcasing its ability to handle tasks requiring precise manipulation and deformable object control. Finally, the RopeThreadingEnv (Thread Manipulation Track) showed a significant \textbf{66.67\%} boost in performance.

These results are summarized in Table \ref{tab:success_rates}, where MedFlamingo PPO consistently outperforms both Baseline PPO and OpenFlamingo PPO across all environments.

\begin{table}[ht]
\centering
\caption{Best Policy Success Rates Across Environments}
\label{tab:success_rates}
\begin{tabular}{lp{1.5cm}p{1.5cm}p{1.5cm}}
\hline
\textbf{Environment}         & \textbf{Baseline PPO (\%)} & \textbf{OpenFlamingo PPO (\%)} & \textbf{MedFlamingo PPO (\%)} \\ \hline
RopeCuttingEnv               & 23                    & 68                         & \textbf{82}              \\
TissueManipulationEnv        & 40                    & 48                         & \textbf{71}              \\
DeflectSpheresEnv            & 7                     & 77                         & \textbf{85}              \\
RopeThreadingEnv             & 57                    & 80                         & \textbf{95}              \\
PickAndPlaceEnv              & 26                    & 62                         & \textbf{80}              \\ \hline
\end{tabular}
\end{table}

\section{Discussion}

We demonstrate a substantial performance improvement achievable through the integration of domain-specific VLMs with RL algorithms for robotic surgical tasks. In particular, MedFlamingo PPO consistently outperformed both the baseline PPO and OpenFlamingo PPO across all evaluated environments, achieving success rates exceeding 70\%. 

\subsection{Impact of Domain-Specific Knowledge}

The superior performance of MedFlamingo PPO compared to OpenFlamingo PPO underscores an importance of domain-specific knowledge in robotic surgical planning and decision-making. While both models share a similar architectural foundation, MedFlamingo PPO's medical domain adaptation possibly enables it to better understand and process task-relevant visual and textual information. For instance, the DeflectSpheresEnv, which involves complex spatial reasoning under constrained motion, saw MedFlamingo PPO achieve a remarkable improvement over Baseline PPO. This improvement may suggest that MedFlamingo's fine-tuning with medical expertise is particularly beneficial for tasks requiring nuanced understanding of spatial relationships and constraints.

By contrast, OpenFlamingo PPO, which lacks domain-specific fine-tuning, demonstrated consistent performance improvements over Baseline PPO but fell short of MedFlamingo PPO's results in all environments. This trend was most evident in the PickAndPlaceEnv, where precise manipulation of a deformable object required specialized control strategies. MedFlamingo PPO achieved a 207.69\% improvement over Baseline PPO in this task, compared to OpenFlamingo PPO's more modest gains. These results align with findings in other domains, where specialized models consistently outperform general-purpose counterparts in complex, high-stakes tasks.

\subsection{Role of Vision-Based Observations}

The exclusive use of endoscopic vision-based observations in this study presents both challenges and opportunities for robotic policy learning. Vision-based observations more closely resemble real-world surgical scenarios but introduce significant complexities in interpreting visual feedback for precise decision-making. Traditional PPO, reliant solely on ResNet-encoded visual features, struggled to achieve task success rates beyond 30\% across most environments. This highlights the limitations of vision-only models in extracting high-level semantic understanding from raw visual data.

MedFlamingo PPO addresses this limitation by leveraging high-level planning tokens generated from task observations and instructions at the start of each episode. These tokens encapsulate task-specific semantic information, reducing the reliance on visual features alone for real-time decision-making. The results suggest that combining vision-based observations with domain-informed planning significantly enhances both the accuracy and efficiency of robotic policies.

\begin{figure}
    \centering
    \includegraphics[width=1\linewidth]{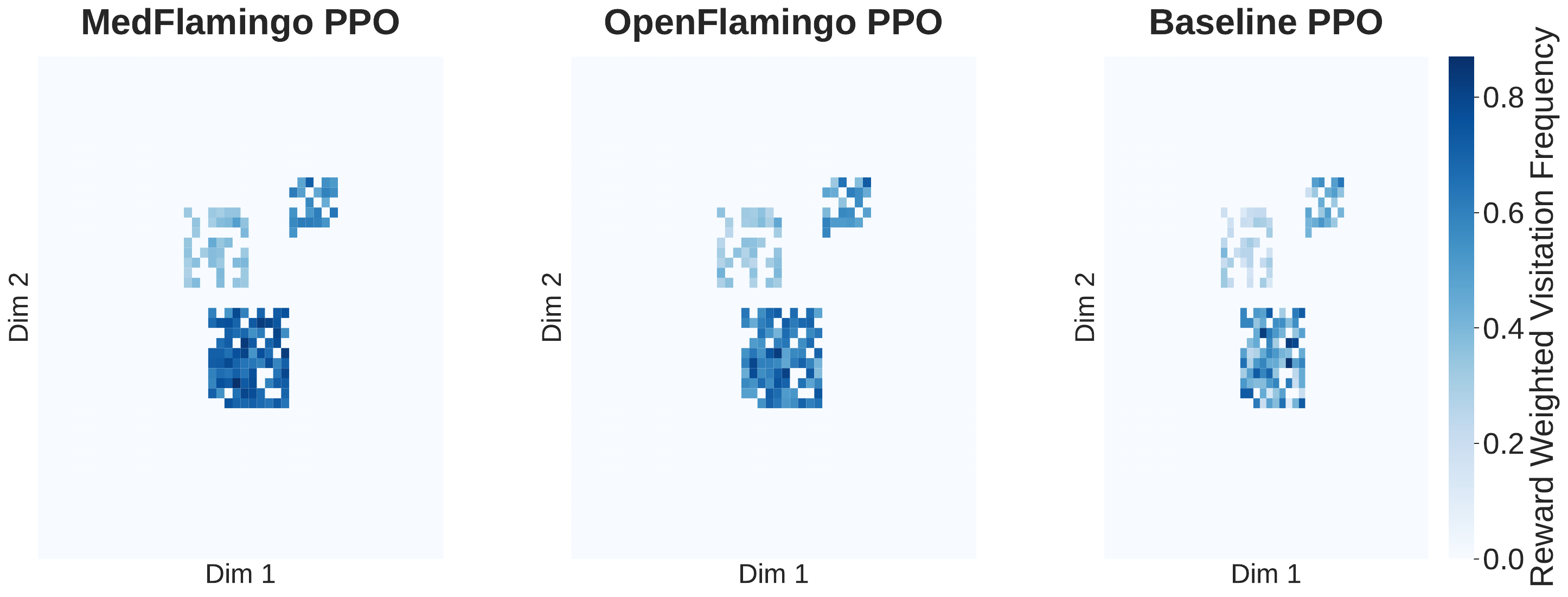}
    \caption{Reward-weighted state visitation heatmaps for MedFlamingo PPO, OpenFlamingo PPO, and Baseline PPO in RopeThreadingEnv. Clustered states capture fine-grained exploration patterns, with visitation frequencies weighted by rewards representing task-relevant regions. MedFlamingo PPO demonstrates efficient, targeted exploration focused on high-reward regions, while OpenFlamingo PPO shows moderate focus with some spillover into less relevant states. Baseline PPO exhibits scattered and inefficient exploration, with sparse activity in key areas.}
    \label{fig:state_visitation}
\end{figure}

Moreover, Heatmap analyses of reward-weighted state visitation frequencies, as shown in Figure \ref{fig:state_visitation}, further illustrate this point. In this analysis, RopeThreadingEnv states were clustered and divided into a grid to capture fine-grained exploration patterns. Each grid cell represents a region in the state space, and the normalized frequency of visits of the agent during training was calculated for all three approaches: MedFlamingo PPO, OpenFlamingo PPO, and Baseline PPO. Each grid cell's visitation frequency is weighted by its corresponding reward value, providing a visualization of exploration efficiency across the task environment.

The heatmaps reveal that MedFlamingo PPO achieved a highly efficient exploration pattern, with concentrated visitation in regions of high reward and minimal activity in low-reward areas. In contrast, OpenFlamingo PPO exhibited a broader and less precise exploration pattern, with moderate focus on high-reward regions but spillover into less relevant areas. Baseline PPO displayed sparse and scattered visitation, failing to adequately prioritize task-relevant states and reflecting inefficient exploration.

\subsection{Efficiency and Generalizability}

One of the key contributions of this work is the optimization of VLM inference for reinforcement learning. By processing task instructions and observations once per episode, MedFlamingo PPO achieves competitive efficiency while maintaining robust performance across diverse environments. This design choice addresses the computational challenges of incorporating large-scale VLMs into real-time robotic applications.

Furthermore, the ability of MedFlamingo PPO to generalize across multiple task categories—spanning dissection, grasping, thread manipulation, spatial reasoning, and deformable object control—demonstrates its versatility. The consistent performance improvements across all environments highlight its potential for broader applications in autonomous surgical systems.

\subsection{Limitations and Future Directions}

While MedFlamingo PPO demonstrates clear advantages, several limitations merit discussion. First, the evaluation was conducted exclusively in simulated environments provided by LapGym. Although these environments are carefully designed to mimic real-world laparoscopic tasks, extending this framework to physical robotic systems remains a critical step for practical adoption. Real-world testing would introduce additional complexities such as hardware dynamics, noise, and variability in surgical conditions.

Second, this study employs a fixed approach to generating high-level planning tokens at the start of each episode. Future research could explore dynamic token updates during policy execution, potentially allowing for greater adaptability to evolving task conditions. Additionally, further ablation studies are needed to isolate and quantify the contributions of MedFlamingo PPO's individual components, such as the VLM, ResNet encoder, and PPO architecture.

Lastly, the reliance on pre-trained VLMs raises questions about their scalability and adaptability to novel surgical tasks beyond the training distribution. Investigating methods for fine-tuning or augmenting VLMs with real-world surgical data could further enhance their applicability and robustness.

\subsection{Implications for Robotic Surgery}

The results of this study have significant implications for the future of autonomous robotic surgery. By integrating domain-specific VLMs like MedFlamingo into reinforcement learning frameworks, robotic systems can achieve higher levels of precision, reliability, and efficiency in complex surgical tasks. This advancement has the potential to reduce surgeon workload, improve patient outcomes, and expand access to high-quality surgical care.

Moreover, the demonstrated success of MedFlamingo PPO highlights the importance of interdisciplinary approaches that combine advancements in computer vision, natural language processing, and robotics. As the field progresses, continued collaboration between medical professionals and AI researchers will be essential to address the unique challenges and opportunities presented by robotic surgery.

\subsection{Conclusion}

In summary, we demonstrate the potential of domain-specific VLMs for robotic policy learning in laparoscopic surgery. MedFlamingo PPO's ability to outperform state-of-the-art baselines across diverse environments underscores the value of integrating medical expertise into reinforcement learning frameworks. Future work should focus on real-world evaluations, dynamic token strategies, and the development of scalable, adaptive models to further advance the field of autonomous robotic surgery.


\bibliographystyle{IEEEtran}
\large

\bibliography{IEEE-conference-template-062824}

\end{document}